\documentclass[journal]{IEEEtran}  
\ifCLASSINFOpdf
\else
\fi
\usepackage{amsmath,graphicx,cite}
\usepackage{subfig}
\usepackage{widetext}
\usepackage{amssymb}
\usepackage{rotating}
\usepackage{xcolor}
\usepackage{hyperref}

\begin{document}
%

\title{Derivation of a Constant Velocity Motion Model for Visual Tracking}
%
%
%

\author{\IEEEauthorblockN{Nathanael L. Baisa*,~\IEEEmembership{Member,~IEEE,}}
\thanks{*Nathanael L. Baisa is with the School of Computing and Communications, Lancaster University, Lancaster, LA1 4WA, UK (e-mail: nathanaellmss@gmail.com).}}
\maketitle

\begin{abstract}

Motion models play a great role in visual tracking applications for predicting the possible locations of objects in the next frame. Unlike target tracking in radar or aerospace domain which considers only points, object tracking in computer vision involves sizes of objects. Constant velocity motion model is the most widely used motion model for visual tracking, however, there is no clear and understandable derivation involving sizes of objects specially for new researchers joining this research field. In this document, we derive the constant velocity motion model that incorporates sizes of objects that, we think, can help the new researchers to adapt to it very quickly.

\end{abstract}


%
\IEEEpeerreviewmaketitle

\section{Introduction} \label{sec:introduction}

Visual tracking is an active research field in computer vision which has got many applications such as intelligent surveillance, autonomous driving, robot navigation, human-computer/robot interaction, augmented reality, medical applications, visual servoing, motion-based recognition, video indexing, etc. Generally, the main components of visual tracking are object detection, appearance modeling, motion modeling and filtering. Tracking-by-detection is the most widely accepted visual tracking paradigm in computer vision community\cite{Nat19}\cite{Baisa2019}.

The Bayesian approach is the main approach for estimating the trajectories of targets as they move in the scene. The Bayes filter has two steps: the prediction step which predicts the target state based on dynamical model and the update step which updates the resulting density using a newly available measurement. Two known implementations of this filter are the Kalman filter and its extended versions~\cite{WelBis06}, and the particle filter (PF)~\cite{AruMasGorCla02}, both for single-target tracking. These filters can be extended to track multiple targets using data association i.e. finding associations between targets and observations.

The single-target tracking task can be modeled using the state and the measurement equations~\cite{WelBis06} which describe the probabilistic dependence between the latent state variable and the observed measurement.

\begin{equation}
    x_k =  \mathbf{f}_k(x_{k-1}, u_{k-1}, w_{k-1})
\label{eqn:state}
\end{equation}
and
\begin{equation}
    z_k =  \mathbf{g}_k(x_k, v_k)
\label{eqn:observation}
\end{equation}
\noindent where $\mathbf{f}_k$ and $\mathbf{g}_k$ are non-linear, time-varying functions, $\{u_{k-1}, k \in \mathbb{N} \}$ is the known control input which is not necessarily available (usually used in robotics), and $\{w_{k-1}, k \in \mathbb{N} \}$ and $\{v_k, k \in \mathbb{N}\}$ are assumed to be independent and identically distributed (i.i.d) stochastic processes. Usually, Eq~(\ref{eqn:state}) is assumed to be a Markov process i.e. state $x_{k-1}$ contains all measurement information $z_{k-1}$ up to time $k-1$. Most of the time, the function $\mathbf{f}_k$ is obtained using a state-space model\footnote{A state-space model is a mathematical model of a physical system that is represented by first-order differential equations or difference equations, or possibly by vector-matrix structures.} (also referred to as the time-domain approach).

The goal of tracking is to estimate the states of targets which can be the positions, velocities and sizes of targets. The state sequence is assumed to be stochastic and, therefore, it is looking for the probability density function (pdf) of the target states. Thus, tracking is to estimate $p_{k|k}(x_k|z_{1:k})$, the $pdf$ of the target being in state $x_k$, given all the measurements $z_k$ up to time $k$, based on Eq~(\ref{eqn:state}) and Eq~(\ref{eqn:observation}). The estimation is accomplished recursively in two steps: prediction and update.

The $\textit{prediction step}$ uses the dynamic model defined in Eq~(\ref{eqn:state}) to obtain the prior $pdf$ using Chapman-Kolmogorov equation given by

\begin{equation}
    p_{k|k-1}(x_k|z_{1:k-1}) =  \int f_{k|k-1}(x_k|x)p_{k-1|k-1}(x|z_{1:k-1})dx
\label{eqn:prediction}
\end{equation}

\noindent with $p_{k-1|k-1}(x_{k-1}|z_{1:k-1})$ known from the previous iteration and the transition density $f_{k|k-1}(x_k|x_{k-1})$ determined by Eq~(\ref{eqn:state}).

The $\textit{update step}$ uses Bayes' rule once the measurement $z_k$ is available to get the posterior pdf

\begin{equation}
    p_{k|k}(x_k|z_{1:k}) =  \frac {g_k(z_k|x_k)p_{k|k-1}(x_k|z_{1:k-1})}{\int g_k(z_k|x)p_{k|k-1}(x|z_{1:k-1})dx}
\label{eqn:update}
\end{equation}
\noindent where the likelihood $g_k(z_k|x_k)$ is determined by Eq~(\ref{eqn:observation}).

Since Eqs~(\ref{eqn:prediction}) and~(\ref{eqn:update}) cannot be solved analytically, under the assumption of linearity for Eqs~(\ref{eqn:state}) and (\ref{eqn:observation}) and Gaussianity of the prior $p_{k-1|k-1}(x_{k-1}|z_{1:k-1})$ and of the two noise sources, $w_{k-1}$ and $v_k$, an optimal solution can be obtained using a Kalman filter~\cite{WelBis06}. If Eqs~(\ref{eqn:state}) and (\ref{eqn:observation}) are mildly non-linear, it can be solved sub-optimally using the extended Kalman filter (EKF)~\cite{WelBis06} and unscented Kalman filter (UKF)~\cite{JulUhl04}. The UKF can handle severe non-linearities with more accuracy than the EKF with the same computational complexity. However, a Gaussian assumption is still made. If the true density is non-Gaussian, none of KF, EKF and UKF can handle this; only the particle filter~\cite{AruMasGorCla02} can manage such scenarios. A review of multi-target filters such as probability hypothesis density (PHD) filter and multiple hypothesis tracking (MHT) is given in~\cite{VoMalBarCorOsbMahVo15} and their implementation is given here\footnote{\url{https://github.com/nathanlem1/MTF-Lib}}. Kalman-filter based multi-sensor data fusion is given in~\cite{GanHar01}\cite{YukXicZhi07}\cite{Hur02}. Multi-target visual tracking by fusing data from overlapping cameras is also given in~\cite{JiaFanXio18}.

\section{Motion Models} \label{sec:MM}

There are a variety of motion models in the literature~\cite{LiJil03} such as Constant Velocity (CV), Constant Acceleration (CA), Constant Turn (CT), Random Walk (RW), etc. When a linear Gaussian system is assumed, the constant velocity motion model can be used with a Kalman filter~\cite{WelBis06}; RW motion model can also be used with the Kalman filter. In fact, you can reduce tracking errors by using a more complex motion models such as CA or CT. To do that, you need to define a different tracking filter such as EKF or UKF.

For visual tracking, the constant velocity motion model is the most widely used motion model since movements of, for instance, pedestrians can be described well enough using this motion model. For filtering simulation~\cite{WelBis06}\cite{NatAnd19} which deals with points, there are derivation of many motion models including for maneuvering targets~\cite{LiJil03}. However, for visual tracking which includes sizes of objects into account while tracking, there is no clearly understandable and explicit derivation for new researchers even though many visual tracking methods~\cite{BaiWal19}\cite{Nat19}\cite{BewGeott16}\cite{WojBewPau17}\cite{baisa2019occlusionrobust}\cite{BaiBhoWal18} use the CV motion model in many forms of target representations such as detection box centre with width and height~\cite{Nat19,Baisa2019}, with area (scale) and aspect ratio~\cite{BewGeott16}, with height and aspect ratio~\cite{WojBewPau17}, etc. and with or without velocities of the sizes of objects. In the following section, we derive the constant velocity motion model for objects represented with width and height in addition to detection box centre as an example and, obviously, the other forms can be derived using the same fashion.

\section{Derivation of a Discrete Constant Velocity Motion Model} \label{sec:CVderivation}

The Kalman filter~\cite{WelBis06} is a closed-form solution of the Bayes filter that assumes a linear Gaussian system. Each target follows a linear Gaussian model:

\begin{equation}
    f_{k|k-1}(x|\zeta) =  \mathcal{N}(x;F_{k-1}\zeta, Q_{k-1})
\label{eqn:linearState1}
\end{equation}
\noindent
\begin{equation}
    g_{k}(z|x) =  \mathcal{N}(z;H_{k} x, R_{k})
\label{eqn:linearObservation1}
\end{equation}
\noindent where $f_{k|k-1}(.|\zeta)$ is the single target state transition probability density at time k given the previous state $\zeta$ and $g_{k}(z|x)$ is the single target likelihood function which defines the probability that $z$ is generated (observed) conditioned on state $x$. $\mathcal{N}(.;m, P)$ denotes a Gaussian density with mean $m$ and covariance $P$; $F_{k-1}$ and $H_k$ are the state transition and measurement matrices, respectively. $Q_{k-1}$ and $R_k$ are the covariance matrices of the process and the measurement noises, respectively. Note that the control input $u_{k-1}$ in Eq~(\ref{eqn:state}) is assumed zero here which is usual in visual tracking. The measurement noise covariance $R_k$ can be measured off-line from sample measurements i.e. from ground truth and detection of training data~\cite{WelBis06} as it indicates detection performance.

Now, our goal is to get the formulation for $F_{k}$, $Q_{k}$, $H_{k}$ and $R_{k}$. Let's assume the detection box centre points to be estimated are denoted by ($x_{b,k}, y_{b,k}$), and the width and height of the detection box in image coordinates to be estimated are represented by $w_{b,k}$ and $h_{b,k}$, respectively, at time $k$. The velocities of the detection box centre points are also denoted by $\dot{x}_{b,k}$ and $\dot{y}_{b,k}$.

\subsubsection{Derivation for $F_{k-1}$ and $Q_{k-1}$} If the velocity of a target is constant i.e. for a uniform rectilinear motion, the process model for the target as it moves from time $k-1$ to time $k$ can be given as

\begin{equation}
\begin{array} {lll}
    x_{b,k} =& x_{b,k-1} + \triangle T \dot{x}_{b,k-1} \\
    \ddot{x}_{b,k-1} =&  0
\end{array}
\label{eqn:constVelocity}
\end{equation}
\noindent where $\ddot{x}_{b,k-1}$ is acceleration and $\triangle T$ is the sampling period. However, the assumption of perfect constant velocity is unrealistic, particularly for real world applications. Therefore, a relaxation is allowed by introducing a piecewise constant white acceleration, thus, a realistic process model can be given as
\begin{equation}
    x_{b,k} = x_{b,k-1} + \triangle T \dot{x}_{b,k-1} + \frac{\triangle T^2}{2} w_{x,k-1}
\label{eqn:constVelocityRealistic}
\end{equation}
\noindent where $\ddot{x}_{b,k-1} = w_{x,k-1}$. This piecewise constant white acceleration can be described by a zero-mean Gaussian white noise as

\begin{equation}
    w_{x,k-1} =  \ddot{x}_{b,k-1} \sim \mathcal{N}(0, \sigma_{{x,k-1}}^2)
\label{eqn:constAcceleration_wk}
\end{equation}
\noindent where $\sigma_{{x,k-1}}^2$ is a variance which controls the level of relaxation of the constant velocity assumption.

Therefore, as a target moves from time $k-1$ to time $k$, the target state evolves as

\begin{equation}
\begin{array} {lll}
x_{b,k} =& x_{b,k-1} + \triangle T \dot{x}_{b,k-1} + \frac{\triangle T^2}{2} w_{x,k-1}, \\  
y_{b,k} =& y_{b,k-1} + \triangle T \dot{y}_{b,k-1} + \frac{\triangle T^2}{2} w_{y,k-1}, \\
\dot{x}_{b,k} =& \dot{x}_{b,k-1} + \triangle T w_{x,k-1},  \\
\dot{y}_{b,k} =& \dot{y}_{b,k-1} + \triangle T w_{y,k-1},  \\
w_{b,k} =& w_{b,k-1} + w_{w,k-1},  \\
h_{b,k} =& h_{b,k-1} + w_{h,k-1},
\end{array}
\label{eqn:StateEqs}
\end{equation}
\noindent where the sampling period $\triangle T$ is defined as the time between frames; usually assumed to be 1 second. $w_{x,k-1}$, $w_{y,k-1}$, $w_{w,k-1}$ and $w_{h,k-1}$ are the process noises corresponding to $x_b$, $y_b$, $w_b$ and $h_b$, respectively. These are basically zero-mean Gaussian white noises.

If we consider the velocity of a bounding box height, for example, the equation for the height and its velocity in Eq~\ref{eqn:StateEqs} is modified as

\begin{equation}
\begin{array} {lll}
h_{b,k} =& h_{b,k-1} + \triangle T \dot{h}_{b,k-1} + \frac{\triangle T^2}{2} w_{h,k-1}, \\  
\dot{h}_{b,k} =& \dot{h}_{b,k-1} + \triangle T w_{h,k-1},  \\
\end{array}
\label{eqn:StateEqh}
\end{equation}
\noindent Similar method applies if we are interested in including the velocity of width of a detection box.

For now we focus on Eq~\ref{eqn:StateEqs} (without width and height velocities) whose state-space model can also be expressed as a vector-matrix representation in the following form

\begin{widetext}
\begin{equation}
\underbrace{\left[ \begin{array}{c}  
x_{b,k} \\ y_{b,k} \\ \dot{x}_{b,k} \\ \dot{y}_{b,k} \\ w_{b,k} \\ h_{b,k}
\end{array} \right]}_{X_k} =
\underbrace{\begin{bmatrix}
1 & 0 & \triangle T & 0  & 0 & 0 \\ 0 & 1 & 0 & \triangle T & 0  & 0 \\ 0 & 0 & 1 & 0  & 0 & 0 \\ 0 & 0 & 0 & 1 & 0  & 0 \\ 0 & 0 & 0 & 0 & 1 & 0 \\ 0 & 0 & 0 & 0 & 0  & 1
\end{bmatrix}}_{F_{k-1}}
\underbrace{\left[ \begin{array}{c}
x_{b,k-1} \\ y_{b,k-1} \\ \dot{x}_{b,k-1} \\ \dot{y}_{b,k-1} \\ w_{b,k-1} \\ h_{b,k-1}
\end{array} \right]}_{X_{k-1}}
+
\underbrace{\left[ \begin{array}{c}
\frac{\triangle T^2}{2} w_{x,k-1} \\ \frac{\triangle T^2}{2} w_{y,k-1} \\ \triangle T w_{x,k-1} \\ \triangle T w_{y,k-1} \\ w_{w,k-1} \\ w_{h,k-1}
\end{array} \right]}_{W_{k-1}}
\label{eqn:StateSpace}
\end{equation}
\end{widetext}
\noindent Eq~(\ref{eqn:StateSpace}) can be expressed as

\begin{equation}
    X_k =  F_{k-1} X_{k-1} + W_{k-1}
\label{eqn:StateEqn}
\end{equation}
\noindent where the value of the state transition matrix $F_{k-1}$ is given in Eq~(\ref{eqn:StateSpace}) and $W_{k-1} \sim \mathcal{N}(0, Q_{k-1})$. Thus, the value of $Q_{k-1}$ can be obtained by taking the covariance of $W_{k-1}$ as

\begin{equation}
    Q_{k-1} =  Cov(W_{k-1}) = E[W_{k-1} W_{k-1}^T]
\label{eqn:StateEqnCov}
\end{equation}
\noindent where $E[W_{k-1} W_{k-1}^T]$ is the expected value or mean of $W_{k-1} W_{k-1}^T$ and $W_{k-1}^T$ is the transpose of $W_{k-1}$. Now, $Q_{k-1}$ can be given as

\begin{widetext}
\begin{equation}
Q_{k-1}
\begin{array} {lll}
=& E\Bigg[\left[ \begin{array}{c}
\frac{\triangle T^2}{2} w_{x,k-1} \\ \frac{\triangle T^2}{2} w_{y,k-1} \\ \triangle T w_{x,k-1} \\ \triangle T w_{y,k-1} \\ w_{w,k-1} \\ w_{h,k-1}
\end{array} \right]
\left[ \begin{array}{c}
\frac{\triangle T^2}{2} w_{x,k-1} ~~ \frac{\triangle T^2}{2} w_{y,k-1} ~~ \triangle T w_{x,k-1} ~~ \triangle T w_{y,k-1} ~~ w_{w,k-1} ~~ w_{h,k-1}
\end{array} \right] \Bigg],  \\

=& \begin{bmatrix}
\frac{\triangle T^4}{4} \sigma_{w_{x}}^2 & 0 & \frac{\triangle T^3}{2} \sigma_{w_{x}}^2 & 0  & 0 & 0 \\
0 & \frac{\triangle T^4}{4} \sigma_{w_{y}}^2 & 0 & \frac{\triangle T^3}{2} \sigma_{w_{y}}^2 & 0  & 0 \\
\frac{\triangle T^3}{2} \sigma_{w_{x}}^2 & 0 & \triangle T^2 \sigma_{w_{x}}^2 & 0  & 0 & 0 \\
0 & \frac{\triangle T^3}{2} \sigma_{w_{y}}^2 & 0 & \triangle T^2 \sigma_{w_{y}}^2 & 0  & 0  \\
0 & 0 & 0 & 0 & \sigma_{w_{w}}^2 & 0 \\
0 & 0 & 0 & 0 & 0  & \sigma_{w_{h}}^2
\end{bmatrix}

\end{array}
\label{eqn:Q_k-1}
\end{equation}
\end{widetext}
\noindent In this derivation for $Q_{k-1}$, two ideas are important:
\begin{itemize}
  \item $E[w_{x,k-1}w_{x,k-1}] = \sigma_{w_{x}}^2$. Similarly, $E[w_{y,k-1}w_{y,k-1}] = \sigma_{w_{y}}^2$, $E[w_{w,k-1}w_{w,k-1}] = \sigma_{w_{w}}^2$ and $E[w_{h,k-1}w_{h,k-1}] = \sigma_{w_{h}}^2$ where $\sigma_{w_{x}}^2$ is the variance ($\sigma_{w_{x}} = \sqrt{\sigma_{w_{x}}^2}$ is the standard deviation).

  \item $E[w_{x,k-1}w_{y,k-1}] = 0$ because there is no correlation between x-axis and y-axis. Similarly, $E[w_{x,k-1}w_{w,k-1}] = 0$, $E[w_{w,k-1}w_{h,k-1}] = 0$, etc.
\end{itemize}

This finalizes the derivation for $F_{k-1}$ and $Q_{k-1}$. It is also important to note that some researchers give the same values for the variances ($\sigma_{w_{x}}^2$, $\sigma_{w_{y}}^2$, $\sigma_{w_{w}}^2$ and $\sigma_{w_{h}}^2$) during their experiment, just for simplification, though it might be important to properly tune them individually.

\subsubsection{Derivation for $H_k$ and $R_k$} We can derive $H_k$ and $R_k$ with the same approach we used for deriving $F_{k-1}$ and $Q_{k-1}$ above. Accordingly, the observation at time $k$ can be given as

\begin{equation}
\begin{array} {lll}
z_{x,k} =& x_{b,k} + v_{x,k}, \\  
z_{y,k} =& y_{b,k} + v_{y,k}, \\
z_{w,k} =& w_{b,k} + v_{w,k},  \\
z_{h,k} =& h_{b,k} + v_{h,k},
\end{array}
\label{eqn:measurementEqs}
\end{equation}
\noindent where ($z_{x,k}, z_{y,k}$) are the centre point of a detection box at time $k$, and $z_{w,k}$ and $z_{h,k}$ are the width and height of a detection box in image coordinates at time $k$. $v_{x,k}$, $v_{y,k}$, $v_{w,k}$ and $v_{h,k}$ are observation noises corresponding to $z_{x,k}$, $z_{y,k}$, $z_{w,k}$ and $z_{h,k}$, respectively, which are basically zero-mean Gaussian white noises. For instance, $v_{x,k} \sim \mathcal{N}(0, \sigma_{z_{x,k}}^2)$. The state-space model in Eq~\ref{eqn:measurementEqs} can also be represented using the following state-space model, particularly using a vector-matrix representation

\begin{equation}
\underbrace{\left[ \begin{array}{c}  
z_{x,k} \\ z_{y,k} \\ z_{w,k} \\ z_{h,k}
\end{array} \right]}_{Z_k} =
\underbrace{\begin{bmatrix}
1 & 0 & 0  & 0 & 0 & 0 \\ 0 & 1 & 0  & 0 & 0 & 0  \\ 0 & 0  & 0 & 0 & 1 & 0 \\ 0 & 0 & 0 & 0 & 0  & 1
\end{bmatrix}}_{H_k}
\underbrace{\left[ \begin{array}{c}
x_{b,k} \\ y_{b,k} \\ \dot{x}_{b,k} \\ \dot{y}_{b,k} \\ w_{b,k} \\ h_{b,k}
\end{array} \right]}_{X_k}
+
\underbrace{\left[ \begin{array}{c}
v_{x,k} \\ v_{y,k} \\ v_{w,k} \\ v_{h,k}
\end{array} \right]}_{V_k}
\label{eqn:StateSpaceZk}
\end{equation}
\noindent Here the $H_k$ is arranged in such a way that it can map the state space into the observation space. Eq~(\ref{eqn:StateSpaceZk}) can be expressed as

\begin{equation}
    Z_k =  H_k X_k + V_k
\label{eqn:measurementEqn}
\end{equation}
\noindent where the value of the measurement matrix $H_k$ is given in Eq~(\ref{eqn:StateSpaceZk}) and $V_k \sim \mathcal{N}(0, R_k)$. Thus, the value of $R_k$ can be obtained by taking the covariance of $V_k$ as

\begin{equation}
    R_k =  Cov(V_k) = E[V_k V_k^T]
\label{eqn:measurementEqnCov}
\end{equation}
\noindent

Now, $R_k$ can be given as

\begin{equation}
R_k
\begin{array} {lll}
=& E\Bigg[\left[ \begin{array}{c}
v_{x,k} \\ v_{y,k} \\ v_{w,k} \\ v_{h,k}
\end{array} \right]
\left[ \begin{array}{c}
v_{x,k} ~~ v_{y,k} ~~ v_{w,k} ~~ v_{h,k}
\end{array} \right] \Bigg],  \\

=& \begin{bmatrix}
\sigma_{v_{x}}^2 & 0  & 0 & 0 \\
0 & \sigma_{v_{y}}^2 & 0  & 0 \\
0  & 0 & \sigma_{v_{w}}^2 & 0 \\
0 & 0 & 0  & \sigma_{v_{h}}^2
\end{bmatrix}

\end{array}
\label{eqn:R_k}
\end{equation}
\noindent where $E[v_{x,k}v_{y,k}] = 0$, $E[v_{x,k}v_{w,k}] = 0$, $E[v_{w,k}v_{h,k}] = 0$ and so on because they are uncorrelated. It is also important to note that some researchers give the same values for the variances ($\sigma_{v_{x}}^2$, $\sigma_{v_{y}}^2$, $\sigma_{v_{w}}^2$ and $\sigma_{v_{h}}^2$) for simplification during their experiment, however, it might be important to properly tune them individually. This finalizes the derivation for $H_k$ and $R_k$.

Similar approach can be used to derive the CV motion model for any target representation you use for visual tracking such as detection box centre with area (scale) and aspect ratio~\cite{BewGeott16}, with height and aspect ratio~\cite{WojBewPau17}, etc. and with or without velocities of the sizes of objects.

\section{Conclusions} \label{Sec:Conclusion}

In this short document, we have derived the constant velocity (CV) motion model for visual tracking applications which include not only detection box centre but also the size of the detection box in a very explicit and understandable way, particularly for detection box centre with width and height of the detection box. We also gave insight for deriving the CV motion model for any target representations for visual tracking tasks. We think that this clear and explicit derivation of the CV motion model can help new visual tracking researchers to adapt to this research field very quickly.

\ifCLASSOPTIONcaptionsoff
  \newpage
\fi

\bibliographystyle{IEEEtran}
\bibliography{egbib}




\end{document}